\documentclass{article}
\usepackage{spconf,amsmath,epsfig}
\usepackage{amsfonts, amssymb}
\usepackage{multicol, multirow}
\usepackage{tabularx}
\usepackage{booktabs}
\usepackage{caption}
\let\OLDthebibliography\thebibliography
\renewcommand\thebibliography[1]{
  \OLDthebibliography{#1}
  \setlength{\parskip}{0pt}
  \setlength{\itemsep}{0pt plus 0.3ex}
}

\begin{document}\sloppy

\def\x{{\mathbf x}}
\def\L{{\cal L}}

\title{CCPA: Long-term Person Re-Identification via Contrastive\\ Clothing and Pose Augmentation}
%
\name{Vuong D. Nguyen$^{*}$ \hspace{1cm} Shishir K. Shah \thanks{$^{*}$Corresponding author: dnguy222@cougarnet.uh.edu}}
\address{Quantitative Imaging Lab, Univeristy of Houston}

\maketitle

\begin{abstract}
Long-term Person Re-Identification (LRe-ID) aims at matching an individual
across cameras after a long period of time, presenting variations in clothing, pose, and viewpoint. In
this work, we propose \textbf{CCPA}: \textbf{C}ontrastive \textbf{C}lothing
and \textbf{P}ose \textbf{A}ugmentation framework for LRe-ID. Beyond appearance, CCPA captures body shape information which is cloth-invariant using a Relation Graph
Attention Network. Training a robust LRe-ID model requires a wide range of clothing variations
and expensive cloth labeling, which is lacked in current LRe-ID datasets.
To address this, we perform clothing and pose transfer across identities
to generate images of more clothing variations and of different persons wearing similar
clothing. The augmented batch of images serve as inputs to our proposed Fine-grained
Contrastive Losses, which not only supervise the Re-ID model to learn discriminative
person embeddings under long-term scenarios but also ensure in-distribution data generation. Results on LRe-ID datasets demonstrate the effectiveness of our CCPA framework.
\end{abstract}
\section{Introduction}
\label{sec:intro}

\begin{figure}[t]
    \centering
    \includegraphics[width=1\columnwidth]{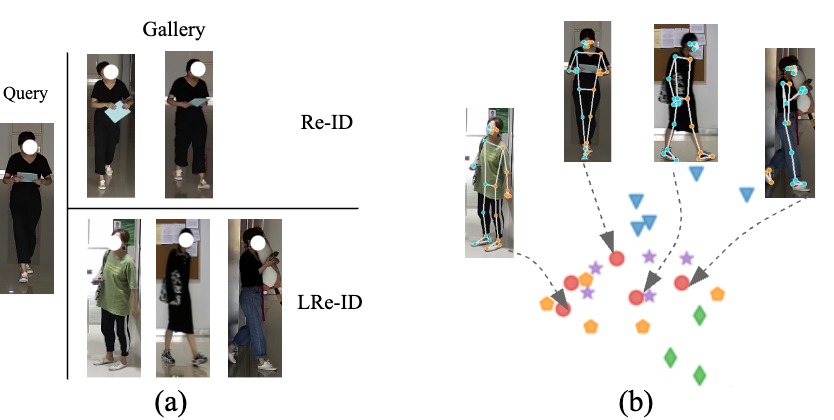}
    \caption{(a) Long-term Re-ID aims to re-identifies the same person under different clothing, viewpoint, illumination, etc.; (b) t-SNE visualization of distribution in the latent space of person embeddings output by the model trained without the proposed losses. Viewpoint variations cause severe ambiguity in learning pose-based shape, leading to large intra-class and small inter-class gap, shown by the widespread of red circles. This forms the motivation of this paper.}
    \label{fig:lreid}
\end{figure}

Person Re-Identification (Re-ID) involves matching the same person
in a non-overlapping camera system. Since the emergence of deep learning, person
Re-ID has been well advanced with plenty of efforts \cite{Sun_2018_PCB,Zhou2019,Xu2018_tra-reid}.
These works assume a simplistic Re-ID scenario where the target person
reappears after a short span of time and space with the same clothing,
pose, and viewpoint. Thus, they suffer severe performance degradation
under long-term scenarios where clothing, pose, and viewpoint have
changed, leading to unreliable appearance. This shortcoming opens a more practical Re-ID problem namely Long-term
Person Re-ID (LRe-ID) (Fig. \ref{fig:lreid}(a)). 

Several methods have been proposed to tackle cloth-changing situations
in LRe-ID, in which two main categories can be observed: single-modality
and multi-modality. Single-modality methods \cite{Gu2022_cal,Cui2023Dcr-reid,Huang2021-clothlabel, Han2023augment, Yang2023cloth-debias}
rely primarily on appearance, face, hairstyle, and cloth labels and
templates from the RGB images to learn person representations. First,
this approach fails under occlusion which makes appearance unobservable.
Second, these texture-based models require large-scale cloth-changing
data with explicit clothing labels and templates as auxiliary labels.
Current LRe-ID datasets only present a small range of clothing variations
due to the difficulty in collecting and labeling such data. Moreover,
clothing labels are ambiguous across identities, thus directly leveraging
manually annotated labels for identity-relevant feature learning is
not beneficial. Multi-modality methods aim to capture cloth-invariant
modalities such as silhouettes \cite{Hong2021shape,jin2022gait},
contour sketches \cite{Yang_2021_prcc,Chen2022shape}, or skeleton-based
pose \cite{qian2020ltcc}, which help provide information beyond appearance
to distinguish individuals. However, viewpoint variations make it
challenging to sufficiently capture fine-grained feature from these
2D modalities for Re-ID as shown in Figure \ref{fig:lreid}(b). 

To this end, we propose \textbf{C}ontrastive \textbf{C}lothing and
\textbf{P}ose \textbf{A}ugmentation (CCPA) framework for LRe-ID. Beyond
appearance, CCPA extracts body shape information from skeleton-based
pose using a Relational Graph Attention Network \cite{busbridge2019rgat}.
We then perform clothing and pose transfer across identities leveraging
appearance and shape. CCPA addresses the lack of clothing variations
in current LRe-ID datasets by synthesizing images of different clothes
for each person. Furthermore, CCPA augments training data by generating
images of different identities wearing similar clothing, which is
a confusing real-world scenario for Re-ID and not explicitly presented
in current LRe-ID datasets. 

Current works that attempt to generate Re-ID data \cite{ge2018poseGAN,khatun2023poseGAN,huang2018GAN}
construct the generation process separately from Re-ID learning, thus
limiting the gain from generated data. Moreover, samples are generated
randomly without identity-related optimization target, leading to
out-of-distribution data which may not be useful for Re-ID. To exploit
the augmentation efficiently, we propose novel
losses which are designed specifically to serve the discriminative
Re-ID learning to better make use of the generated data, while ensuring
a robust generation module. The augmented batch of images is constrastively
sampled as inputs for: (1) the Fine-grained Contrastive Clothing-aware
Loss (FCCL), which mitigates large intra-ID and small inter-ID clothing
variations, and (2) Fine-grained Contrastive Viewpoint-aware Loss
(FCVL), which enforces sufficient penalization on the images of different
IDs under same viewpoint and same ID under different viewpoints. FCCL
helps to learn a robust appearance encoder under clothing confusion,
while FCVL enhances discriminative power of shape representations
under viewpoint changes. 

Our contributions are summarized as follows: (1) we propose CCPA,
a framework for LRe-ID which can extract both identity-relevant and
cloth-irrelevant features; (2) we tackle two major challenges for
LRe-ID including clothing changes and viewpoint variations, in which
clothing and pose transfer is performed to augment training batch
as inputs for the proposed fine-grained contrastive losses. We not
only enhance a more robust representation learning but also address
the lack of cloth-changing data; (3) extensive experiments on LRe-ID
datasets show that we achieve state-of-the-art Re-ID performance in
real-world scenarios.

\section{Related Work}

\subsection{Long-term Person Re-ID}

Traditional Person Re-ID which assumes short-term scenarios has been
well advanced using deep learning models for representation learning
under both supervised \cite{Sun_2018_PCB,Zhou2019,Xu2018_tra-reid} and
unsupervised \cite{li2018illisvid,Chen_2021_CVPR_imggen_gcl,Khaldi_2024_WACV}
setting. However, these appearance-based methods suffer performance
degradation in long-term environment due to significant clothing changes. 

\noindent
\textbf{Cloth-Changing Person Re-ID} (CCRe-ID). With the release of
CCRe-ID datasets \cite{qian2020ltcc,Yang_2021_prcc}, several
CCRe-ID methods have been proposed. Clothing status and
templates are mined as pseudo-labels in \cite{Huang2021-clothlabel,Han2023augment},
which fail when meeting new clothing variations under incremental
data scenario. Texture-based methods \cite{Gu2022_cal,Cui2023Dcr-reid}
rely heavily on visibility of faces and body parts, which fail under
occlusion. Cloth-invariant body shape features are captured from silhouettes
\cite{Hong2021shape,jin2022gait}, contour sketches \cite{Yang_2021_prcc,Chen2022shape},
or skeleton-based pose \cite{qian2020ltcc,Nguyen_2024_ASGL} to assist CCRe-ID. However,
viewpoint variations significantly limit the representation learning
from these 2D modalities.

\noindent
\textbf{Viewpoint-aware Person Re-ID}. View-transformed feature extractors are designed in \cite{Sarfraz2017viewpoint,ZHENG202119viewpoint}
to attend to frontal viewpoint which is the most informative, which
limits the applicability under clothing changes. Zhu \textit{et al.} \cite{zhu2020viewpoint}
proposed to use a viewpoint-aware hyper-sphere to cluster identities,
while a viewpoint-aware feature fusion model is designed in \cite{Ai2020viewpoint}.
However, these methods are not robust when
different IDs wearing similar clothing under the same viewpoint.
3D shape is leveraged in \cite{Chen2021,Nguyen_2024_SEMI}, however,
capturing 3D human models requires heavy training with expensive 3D
data. 

\subsection{Data Augmentation-based Re-ID}

Data augmentation has been applied to improve the feature learning
ability of Re-ID models. Besides traditional methods such as horizontal
clipping or random erasing, GANs are widely used for image generation
\cite{ge2018poseGAN,khatun2023poseGAN}. These works
leverage body pose to condition the generation of pedestrian images
in various poses, which aims to enhance robustness of models under
pose variations. Synthesizing data under different camera styles has
also been exploited \cite{zhong2018camGAN,zhong2019camGAN}. However,
these works have not considered long-term scenarios which is crucial
for real-world Re-ID. 

\section{Methodology}

\subsection{The Proposed Framework }

\begin{figure*}
    \centering
    \includegraphics[width=0.9\textwidth]{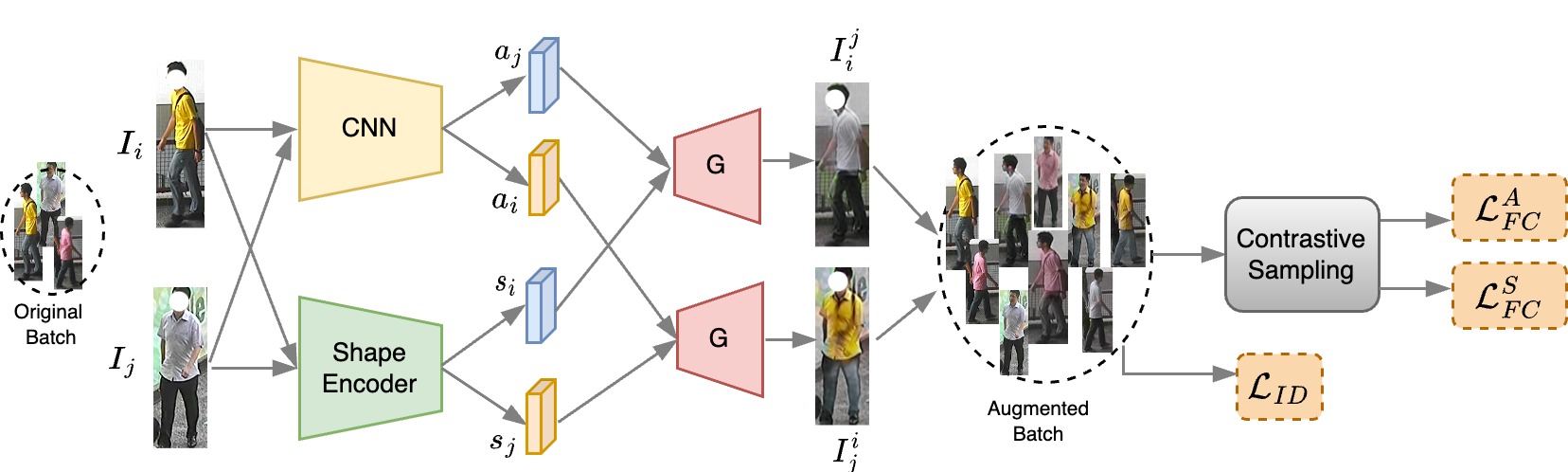}
    \caption{Overview of the proposed CCPA framework. From the original batch, for every pair of images, clothing and pose transfer is performed using an appearance encoder $E^A$ (a CNN), a shape encoder $E^S$ and a decoder $G$ to form the augmented batch. Contrastive sampling is performed on the augmented batch for inputs to the proposed fine-grained contrastive losses, which drive model training along with an identification loss. (Best viewed in color)}
    \label{fig:baseline}
\end{figure*}

An overview of our proposed framework is given in Fig. \ref{fig:baseline}. Given
an original batch of $N$ images $X=\{I_{i}\}_{i=1}^{N}$, denote
the identity label set as $\{y_{i}\}_{i=1}^{N}$ (note that $y_{i}\ne y_{j}$
for $i,j=1,...,N$). The framework comprises an appearance encoder
$E^{A}$ which is a CNN and a shape encoder $E^{S}$. For every pair
of images, we perform clothing and pose transfer to obtain an
augmented batch. Specifically, a pair of images $[I_{i},I_{j}]\in X$
is sent to $E^{A}$ and $E^{S}$ which output appearance embeddings
$a_{i}=E^{A}(I_{i}),a_{j}=E^{A}(I_{j})$ and shape embeddings $s_{i}=E^{S}(I_{i}),s_{j}=E^{S}(I_{j})$.
Details about shape representation learning will be discussed later
in Sec. \ref{sec:shape}. Then, a decoder $G$ takes in and swaps appearance and shape
embeddings to generate new pedestrian images, i.e. $I_{i}^{j}=G(s_{i},a_{j}),I_{j}^{i}=G(s_{j},a_{i})$.
The identity label of the synthesized image corresponds to the input
image that gives shape embedding, i.e. $y_{I_{i}^{j}}=y_{I_{i}}$,
since we only apply new clothing from $a_{j}$ on the old person.
The augmented batch $X'$ containing $N^{2}$ images is then contrastively
sampled as inputs for the proposed Fine-grained Contrastive Losses
$\mathcal{L}_{FC}^{A}$ and $\mathcal{L}_{FC}^{S}$. An identification
loss $\mathcal{L}_{ID}$ based on cross entropy loss is also employed
as classification loss of the model. Input of $\mathcal{L}_{ID}$
is the person representation $f_{i}$ obtained by concatenating appearance
and shape embeddings, i.e. $f_{i}=[a_{i},s_{i}]$, along with identity label $y_i$. The framework is trained by the
total loss $\mathcal{L}$ formulated as:
\begin{equation}
\mathcal{L}=\mathcal{L}_{FC}^{A}+\mathcal{L}_{FC}^{S}+\mathcal{L}_{ID}
\end{equation}

\begin{figure}
    \centering
    \includegraphics[width=0.98\columnwidth]{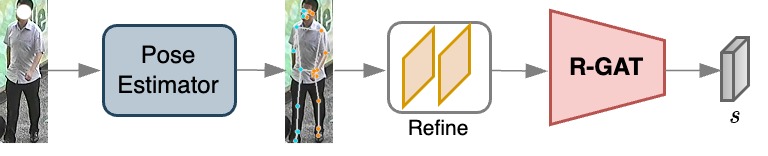}
    \caption{Architecture of Shape Encoder, which comprises of a pose estimator, a refinement network and a R-GAT.}
    \label{fig:shape}
\end{figure}
\subsection{Shape Representation Learning\label{sec:shape}} 

To tackle long-term Re-ID scenarios, skeleton-based shape remains
a competitive modality since it is invariant to clothing variations.
The architecture of the Shape Encoder is illustrated in Fig. \ref{fig:shape}.
Given an image $I\in X$, pose is estimated using OpenPose \cite{cao2017openpose}
which outputs a set of $m$ joint nodes $\mathbf{J}=\{\mathbf{j}_{i}\}_{i=1}^{m}$.
These nodes are then transformed into higher-dimensional feature vectors
$\mathbf{J}'=\{\mathbf{j}'_{i}\}_{i=1}^{m},\mathbf{j}'_{i}\in\mathbb{R}^{F}$
by a refinement network consisting of several fully-connected layers.
Intuitively, to represent body shape, we need to capture the
local and global relationships among nodes. To achieve this, we employ
Relational Graph Attention Network (R-GAT) \cite{busbridge2019rgat}.
R-GAT differs from GCN by incorporating attention mechanism into
graph operation to account for cross-node importance, while compared
to GAT \cite{velivckovic2017gat}, R-GAT also amplifies different
levels of relations among bones, which is crucial for capturing a
discriminative shape when some body parts are invisible due to occlusion
or extreme viewpoints. 

Input of R-GAT is a graph constructed from $\mathbf{J}'$
with $m$ nodes and $k$ relations ($k$ edges/bones). Let
$\mathbf{J}'=[\mathbf{j}'_{1},...,\mathbf{j}'_{m}]\in\mathbb{R}^{m\times F}$
be the input feature matrix, the intermediate representation matrix
under a relation $\mathbf{r}$ is computed as:
\begin{equation}
\mathbf{H^{(\mathbf{r})}}=\mathbf{J}'W^{(\mathbf{r})}\in\mathbb{R}^{m\times F'},
\end{equation}
where $\mathbf{H^{(\mathbf{r})}}=[\mathbf{h}_{1}^{(\mathbf{r})},...,\mathbf{h}_{m}^{(\mathbf{r})}]$
and $W^{(\mathbf{r})}\in\mathbb{R}^{F\times F'}$ are learnable parameters.
For each node $\mathbf{h}_{i}^{(\mathbf{r})}\in\mathbf{H}$, denote
the indices set of its neighbors as $\mathcal{N}_{i}^{(\mathbf{r})}$,
$\mathbf{h}_{i}^{(\mathbf{r})}$ is updated after a R-GAT layer based
on a weighted sum over the nodes in its neighborhood and over all
the relations as follows:

\begin{equation}
\mathbf{\hat{h}}_{i}=\sigma(\sum_{\mathbf{r}}\sum_{j\in\mathcal{N}_{i}^{(\mathbf{r})}}\alpha_{i,j}^{(\mathbf{r})}\mathbf{\hat{h}}_{j}^{(\mathbf{r})}),\,\,\,\alpha_{i,j}^{(\mathbf{r})}=\text{SM}\left(\mathbf{h}_{i}^{(\mathbf{r})}\mathbf{h}_{j}^{(\mathbf{r})}\right),
\end{equation}
where $\alpha_{i,j}^{(\mathbf{r})}$ is attention score between the
$i^{th}$ and $j^{th}$ nodes, $\sigma$ is sigmoid activation function,
and $\text{SM}$ denotes softmax. The graph is finally aggregated
to obtain shape embedding $s$ by computing the mean of node representations
as: 
\begin{equation}
s=\frac{1}{m}\sum_{i=1}^{m}\mathbf{\hat{h}}_{i}.
\end{equation}

\subsection{Fine-grained Contrastive Losses}

Given the augmented batch $X'$, denote the identity label set as
$\{y_{i}\}_{i=1}^{N^{2}}$, clothing label set as $\{c_{i}\}_{i=1}^{N^{2}}$,
and viewpoint label set as $\{v_{i}\}_{i=1}^{N^{2}}$. To enhance the robustness of appearance encoder under cloth-confusing
scenarios, we propose the \textbf{Fine-grained Contrastive Clothing-aware Loss} (FCCL),
whose inputs are sampled as follows: for an
appearance embedding $a_{i}$ of $I_{i}\in X'$ as anchor, the set
of positive samples (denoted as $P_{i}^{A+}$) consists of images of the same identity, while the set of negative
samples (denoted as $P_{i}^{A-}$) 
consists of images of different identities. Since in the original batch
we only sample one image per identity, positive samples in $P_{i}^{A+}$
have a wide range of clothing variations transferred from other identities,
while negative samples include persons in similar clothing as anchor.
Therefore, the proposed FCCL helps enforce a large penalization on
small inter-class and large intra-class difference caused by ambiguity
in clothing variations. FCCL is formulated as: 

\begin{equation}\label{eq:loss_app}
\mathcal{L}_{FC}^{A}=-\sum_{i=1}^{N^{2}}\gamma_{1}\log\frac{\sum_{j\in P_{i}^{A+}}\eta(a_{i},a_{j})}{\sum_{k\in P_{i}^{A-}}\eta(a_{i},a_{k})},
\end{equation}
where $P_{i}^{A+}=\left\{ j\,|\,I_{j}\in X'\,|\,y_{j}=y_{i},c_{j}\ne c_{i}\right\} $,
$P_{i}^{A-}=\left\{ k\,|\,I_{k}\in X'\,|\,y_{k}\ne y_{i}\right\} $,
$\eta(\cdot,\cdot)$ is a distance-to-distribution function, and $\gamma_{1}$ is a scalar controlling the scale
of $\mathcal{L}_{FC}^{A}$. 

To mitigate the ambiguity in shape caused by viewpoint variations
as shown in Fig. \ref{fig:lreid}, we propose \textbf{Fine-grained Contrastive Viewpoint-aware
Loss} (FCVL) $\mathcal{L}_{FC}^{S}$, whose inputs are sampled as
follows: for a shape embedding $s_{i}$ of $I_{i}\in X'$ as anchor,
images of the \textbf{same identity} but \textbf{different viewpoints}
are chosen as positive samples (denoted as $P_{i}^{S+})$, while images
of \textbf{different identities} under the \textbf{same viewpoint
}are negative samples (denoted as\textbf{ }$P_{i}^{S-})$. Specifically:
\begin{equation}\label{eq:loss_shape}
\mathcal{L}_{FC}^{S}=-\sum_{i=1}^{N^{2}}\gamma_{2}\log\frac{\sum_{j\in P_{i}^{S+}}\eta(a_{i},a_{j})}{\sum_{k\in P_{i}^{S-}}\eta(a_{i},a_{k})},
\end{equation}
where $P_{i}^{S+}=\left\{ j\,|\,I_{j}\in X'\,|\,y_{j}=y_{i},v_{j}\ne v_{i}\right\} $,
$P_{i}^{S-}=\left\{ k\,|\,I_{k}\in X'\,|\,y_{k}\ne y_{i},v_{k}=v_{i}\right\} $,
and $\gamma_{2}$ is a scalar controlling the scale
of $\mathcal{L}_{FC}^{S}$.

The distance-to-distribution function $\eta$ is a key component in
Eq. \ref{eq:loss_app} and \ref{eq:loss_shape}. In this work, we formulate $\eta$ as an exponential
function as follows: $\eta(\cdot,\cdot)=e^{-d(\cdot,\cdot)}$ where
$d(\cdot,\cdot)$ denotes euclidean distance. The exponentially sensitive
penalization enforces the model to pull embeddings of the same identities
closer while pushing that of different identities farther, resulting
in a well-separable latent space. Simultaneously, it ensures that the distribution
of the generated samples is close to original images.

\section{Experiments}

\subsection{Experimental Setup}

\textbf{Datasets}. To validate the performance of our framework,
we use two LRe-ID datasets: LTCC \cite{qian2020ltcc} and PRCC \cite{Yang_2021_prcc}.
LTCC consists of 17,119 images from 152 identities captured under
12 camera views, while PRCC contains 33,698 images of 221 identities
captured by 3 cameras. LTCC data presents large clothing and
viewpoint variations, illumination, and occlusion. On the other hand,
persons in PRCC are mostly captured in frontal viewpoint with good
lighting condition, and each identity only has from 2 to 5 clothing
variations. Thus, LTCC mimics a more challenging Re-ID environment.

\noindent
\textbf{Evaluation Protocols}. CMC at rank k (rank-k accuracy) and
mean Average Precision (mAP) are used as evaluation metrics. Two evaluation
settings are set up. First is Cloth-Changing (CC) where only cloth-changing
samples are used for testing. Second is Standard where for LTCC, both
cloth-consistent and cloth-changing samples form the query and gallery
sets, while for PRCC, due to its nature, the test sets only contain
cloth-consistent samples.

\noindent
\textbf{Implementation Details}. Viewpoint is estimated using MEBOW
\cite{Wu2020MEBOW}. For appearance encoder $E^{A}$, we adopt ResNet-50
\cite{He_2016_CVPR} with pretrained weights on ImageNet \cite{Deng2009imagenet}.
For shape encoding, pose is estimated in COCO format using OpenPose
\cite{cao2017openpose}. The refinement network consists of 2 fully-connected
layers of size {[}512, 2048{]}, while R-GAT consists of 2 layers.
Generator G has four residual blocks and four convolutional layers
and it is initialized with pretrained weights from \cite{zheng2019_imggen_dgnet}.
We set $\gamma_{1}=0.7$ and $\gamma_{2}=0.3$. Input images are resized
to $256\times128$ for training. Each original batch is sampled with
8 images from 8 identities, giving an augmented batch size of 64.
The framework is trained for 90 epochs using Adam optimizer with an
initial learning rate  $lr=0.0005$, momentum of 0.9, and weight decay of 0.01. $lr$ is reduced by a factor of 0.1
after every 30 epochs. 

\subsection{Results}

\begin{table*}
\small 
\begin{centering}
\begin{tabular}{c|c|cc|cc|cc|cc}
\hline 
\multirow{3}{*}{Methods} & \multirow{3}{*}{Modality} & \multicolumn{4}{c|}{LTCC} & \multicolumn{4}{c}{PRCC}\tabularnewline
\cline{3-10} \cline{4-10} \cline{5-10} \cline{6-10} \cline{7-10} \cline{8-10} \cline{9-10} \cline{10-10} 
 &  & \multicolumn{2}{c|}{CC} & \multicolumn{2}{c|}{Standard} & \multicolumn{2}{c|}{CC} & \multicolumn{2}{c}{Standard}\tabularnewline
\cline{3-10} \cline{4-10} \cline{5-10} \cline{6-10} \cline{7-10} \cline{8-10} \cline{9-10} \cline{10-10} 
 &  & R-1 & mAP & R-1 & mAP & R-1 & mAP & R-1 & mAP\tabularnewline
\hline 
\hline 
PCB \cite{Sun_2018_PCB} & RGB & 23.5 & 10.0 & 65.1 & 30.6 & 41.8 & 38.7 & \textbf{99.8} & 97.0\tabularnewline

RCSANet \cite{Huang2021-clothlabel} & RGB & - & - & - & - & 50.2 & 48.6 & \underline{100} & 97.2\tabularnewline


CAL \cite{Gu2022_cal} & RGB & 40.1 & 18.1 & 74.2 & 40.8 & 55.2 & 55.8 & \underline{100} & \underline{99.8}\tabularnewline

CCFA \cite{Han2023augment} & RGB & \textbf{45.3} & \textbf{22.1} & 75.8 & \underline{42.5} & \textbf{61.2} & \underline{58.4} & 99.6 & 98.7\tabularnewline
\hline 
PRCC-contour \cite{Yang_2021_prcc} & RGB + sketch & - & - & - & - & 34.4 & - & 64.2 & -\tabularnewline

CESD \cite{qian2020ltcc} & RGB + pose & 26.2 & 12.4 & 71.4 & 34.3 & -  & - & - & -\tabularnewline

GI-ReID \cite{jin2022gait} & RGB + sil & 23.7 & 10.4 & 63.2 & 29.4 & 33.3 & - & 80.0 & -\tabularnewline

3DSL \cite{Chen2021} & RGB + pose + sil + 3D & 31.2 & 14.8 & - & - & 51.3 & - & - & -\tabularnewline

FSAM \cite{Hong2021shape} & RGB + pose + sil & 38.5 & 16.2 & 73.2 & 35.4 & 54.5 & - & 98.8 & -\tabularnewline



AIM \cite{Yang2023cloth-debias} & RGB + gray & 40.6 & 19.1 & 76.3  & 41.1 & 57.9 & \textbf{58.3 } & \underline{100} & \underline{99.8}\tabularnewline

CVSL \cite{Nguyen_2024_CVSL} & RGB + pose & 44.5 & 21.3  & \textbf{76.4} & 41.9 & 57.5 & 56.9 & 97.5 & 99.1\tabularnewline
\hline 
\textbf{CCPA (Ours)} & RGB + pose & \underline{46.1} & \underline{22.9} & \underline{76.9} & \textbf{42.4} & \underline{61.7} & \underline{58.4} & \underline{100} & \textbf{99.6 }\tabularnewline
\hline 
\end{tabular}
\par\end{centering}
\caption{Quantitative comparison on LTCC and PRCC.
Best results are \underline{underlined}, while second-best results are in \textbf{bold}. } \label{tab:results}
\end{table*}

In Tab. \ref{tab:results}, we provide a quantitative comparison in
results between our method and a state-of-the-art standard Re-ID method
(i.e. PCB \cite{Sun_2018_PCB}) and LRe-ID methods (i.e. the remaining)
on LTCC and PRCC. It can be seen that clothing changes cause relatively
inferior performance to the standard Re-ID method since appearance
is no longer reliable. Among LRe-ID methods, a majority of them \cite{Yang_2021_prcc,qian2020ltcc,jin2022gait,Chen2021,Hong2021shape,li2020shape,wang2022camc,Yang2023cloth-debias,Nguyen_2024_CVSL}
resort to cloth-invariant modalities such as sketch, silhouette, or
pose, with results showing that this approach is effective in mitigating
the influence of clothing changes. Although recent texture-based methods
\cite{Gu2022_cal,Han2023augment} achieve comparable performance,
they rely heavily on the availability of large clothing variations
with expensive manual cloth labeling to be aware of clothing changes.
The influence of viewpoint changes has also been ignored. Moreover,
the discriminative fine-grained information identity information under
similar-clothing scenarios has not been adequately mined. Our CCPA framework achieves
46.1/22.9\% Rank-1/mAP accuracy on LTCC and 61.7/58.4\% Rank-1/mAP
accuracy on PRCC. By addressing
these issues, overall, we outperform previous methods in cloth-changing
environment on both datasets. 

\subsection{Ablation Study}

To further validate the effectiveness of our proposed method, we perform
comprehensive ablation studies on LTCC and PRCC of: Clothing and Pose Augmentation
(CPA), the proposed Fine-grained Contrastive Losses, and the Relational
Graph Attention Network.

\begin{table}
\small 
\begin{centering}
\begin{tabular}{c|c|cc|cc|cc}
\hline 
\multirow{2}{*}{Method} & \multirow{2}{*}{CPA} & \multirow{2}{*}{$\mathcal{L}_{FC}^{A}$} & \multirow{2}{*}{$\mathcal{L}_{FC}^{S}$} & \multicolumn{2}{c|}{LTCC} & \multicolumn{2}{c}{PRCC}\tabularnewline
\cline{5-8} \cline{6-8} \cline{7-8} \cline{8-8} 
 &  &  &  & R-1 & mAP & R-1 & mAP\tabularnewline
\hline 
\hline 
1 & \checkmark & - & - & 42.4 & 19.2 & 56.1 & 54.3\tabularnewline

2 & \checkmark & \checkmark & - & 44.6 & 20.9 & 59.1 & 56.8\tabularnewline

3 & \checkmark & - & \checkmark & 44.0 & 20.6 & 58.8 & 56.2\tabularnewline
\hline 
4 & - & \checkmark & \checkmark & 44.3 & 20.8 & 59.2 & 57.0\tabularnewline
\hline 
CCPA & \checkmark & \checkmark & \checkmark & \underline{46.1} & \underline{22.9} & \underline{61.7} & \underline{58.4}\tabularnewline
\hline 
\end{tabular}
\par\end{centering}
\caption{Ablation study of Clothing and Pose Augmentation
(CPA) and the proposed Fine-grained Contrastive Losses on LTCC and
PRCC in cloth-changing setting.} \label{tab:Ablation-study}
\end{table}

\noindent
\textbf{Clothing and Pose Augmentation.}
From Tab. \ref{tab:Ablation-study}, it can be seen that CPA significantly
improves Re-ID performance on both datasets (method 4 vs CCPA). For
example, Rank-1 accuracy is boosted by 1.8/2.5\% on LTCC/PRCC. This
demonstrates the rationality and effectiveness of our proposed approach
to address the lack of clothing variations and labels in current LRe-ID
datasets. CPA leads to better generalizability by exposing the model
to similar-clothing situations, shown by an improvement of 1.1/1.4\%
in mAP on LTCC/PRCC.

\noindent
\textbf{Fine-grained Contrastive Losses}
Tab. \ref{tab:Ablation-study} demonstrates that overall, with the
proposed Fine-grained Contrastive Losses, Re-ID performance is remarkably
improved on both datasets (method 1 vs CCPA) with an increase of 3.7/3.7\%
in Rank-1/mAP on LTCC and 5.6/4.1\% in Rank-1/mAP on PRCC. We also
run experiments with adding one loss and excluding the other to explore
the contribution of each loss (method 2 and method 3). By leveraging contrastive learning, the clothing-aware loss $\mathcal{L}_{FC}^{A}$
effectively guides the model to distinguish identities under
clothing variations and similar clothing, shown by a boost of 2.2/1.7\%
in Rank-1/mAP on LTCC. The viewpoint-aware loss $\mathcal{L}_{FC}^{S}$ enhances the model's performance by learning shape embeddings
with high discriminative power under viewpoint changes.

\begin{table}
\small 
\begin{centering}
\begin{tabular}{c|cc|cc}
\hline 
\multirow{2}{*}{Method} & \multicolumn{2}{c|}{LTCC} & \multicolumn{2}{c}{PRCC}\tabularnewline
\cline{2-5} \cline{3-5} \cline{4-5} \cline{5-5} 
 & R-1 & mAP & R-1 & mAP\tabularnewline
\hline 
\hline 
GCN & 44.9 & 22.0 & 59.8 & 56.1\tabularnewline

GAT & 45.3 & 22.3 & 60.4 & 57.1\tabularnewline
\hline 
R-GAT & \underline{46.1} & \underline{22.9} & \underline{61.7} & \underline{58.4}\tabularnewline
\hline 
\end{tabular}
\par\end{centering}
\caption{Ablation study of R-GAT on LTCC and PRCC in cloth-changing setting.} \label{tab:rgat}
\end{table}

\noindent
\textbf{Relational Graph Attention Network}
In Tab. \ref{tab:rgat}, we show the effectiveness of using R-GAT
in encoding shape representation from the skeleton-based graph. R-GAT
clearly shows superiority over GAT and GCN. For example, R-GAT outperforms
GAT by 0.8/0.6\% and GCN by 1.2/0.9\% in Rank-1/mAP on LTCC. This
indicates that besides capturing relationship among joints, it is
beneficial to capture the relationship among bones for a robust shape
embedding.

\section{Conclusion}
In this work, we propose CCPA, a novel framework for LRe-ID. CCPA extracts simultaneously appearance and cloth-invariant pose-based shape, then leverages these modalities to transfer clothing and pose across identities for cloth-changing data generation. The augmented data serves as inputs to the Fine-grained Contrastive Losses, which guides the model to learn a well separable latent space of person embeddings under cloth-confusing and viewpoint-changing scenarios. Extensive experiments and ablation studies on LRe-ID datasets validate the effectiveness of our proposed framework.
\small{
\bibliographystyle{IEEEbib}
\bibliography{main}
}

\end{document}